\documentclass{Interspeech2024}
\usepackage{multirow}
\usepackage{subcaption}
\usepackage{caption}
\usepackage[table,xcdraw]{xcolor}
\usepackage[flushleft]{threeparttable}
\interspeechcameraready

\title{A Comprehensive Solution to Connect Speech Encoder and Large Language Model for ASR}

\name[affiliation={}]{Van Tung}{Pham}
\name[affiliation={}]{Yist}{Lin}
\name[affiliation={}]{Tao}{Han}
\name[affiliation={}]{Wei}{Li}
\name[affiliation={}]{Jun}{Zhang}
\name[affiliation={}]{Lu}{Lu}
\name[affiliation={}]{Yuxuan}{Wang}

\address{
  ByteDance}
\email{\{van.pham,yist.lin0,tao.han,liwei.speech,zhangjun.jarry,lulu.0314,wangyuxuan.11\}@bytedance.com}


\begin{document}

\maketitle

\begin{abstract}
    
Recent works have shown promising results in connecting speech encoders to large language models (LLMs) for speech recognition. However, several limitations persist, including limited fine-tuning options, a lack of mechanisms to enforce speech-text alignment, and high insertion errors especially in domain mismatch conditions. This paper presents a comprehensive solution to address these issues. We begin by investigating more thoughtful fine-tuning schemes. Next, we propose a matching loss to enhance alignment between modalities. Finally, we explore training and inference methods to mitigate high insertion errors. Experimental results on the Librispeech corpus demonstrate that partially fine-tuning the encoder and LLM using parameter-efficient methods, such as LoRA, is the most cost-effective approach. Additionally, the matching loss improves modality alignment, enhancing performance. The proposed training and inference methods significantly reduce insertion errors.

\end{abstract}

\section{Introduction}
Recently, large language models (LLM), characterised by their billions of parameters and training on massive datasets, have demonstrated emergent abilities to address various tasks in natural language processing field. Meanwhile, foundation models also advanced state-of-the-art in other research fields such as speech processing \cite{hubert,zhang2023google} and computer vision \cite{dosovitskiy2021image,blip2}. Hence, it is natural to develop effective approaches that unify foundation models from different modalities to build strong speech and visual understanding ability. In this work, we aim to connect a speech foundation model to an LLM for speech recognition. 

The architecture of LLM-based ASR generally consists of three components: a speech encoder, an adapter, and an LLM. Recent works in LLM-based ASR examined techniques to compress the output of speech encoder \cite{Wu2023OnDA,Fathullah2023,slm}, design different adapters \cite{slm,Yu2023ConnectingSE,ling2023adapting}, fine-tune the LLM partially \cite{Fathullah2023} or keep it unchanged \cite{slm}. Despite these efforts, several issues still persist. Firstly, existing works only fine-tuned certain modules with specific configurations, which might exclude many efficient settings. Secondly, there is no method to explicitly force the representations generated by the adapter to be in a similar space, i.e. aligned, to those of LLM embeddings, although \cite{Fathullah2023} shown that these representations are aligned to some extent after ASR fine-tuning. Finally, we observe very high insertion errors when the model is trained with limited training data or when test sets are mismatched with the training set - a problem that has not been studied before. 

In this work, we present a comprehensive solution
to address aforementioned issues. We first perform an empirical study about more  thoughtful fine-tuning settings, each corresponding to different fine-tune options for each module, to identify efficient settings. 
We then propose a matching loss on top of a cross attention to explicitly force the two modalities to align with each other, leading to better ASR performance. Finally, we propose several training and inference strategies to address the high insertion problem. Specifically, for inference, we apply n-gram non-repetition and length penalty constraints during beam search decoding. In terms of training, we propose using a non-speech corpus with empty transcript for fine-tuning and data augmentations based on speed and volume perturbations. 

The paper is organized as follows. Section 2 presents related works. In Section 3, we describe our proposed solution, followed by experimental setup in Section 4. Section 5 presents experimental results and analysis, then we conclude our work in Section 6. 

\section{Related works}
Recently published works have extended LLMs with the ability to ingest other modalities, such as audio \cite{Wu2023OnDA,Fathullah2023,slm,Yu2023ConnectingSE,speechGPT,audioPaLM} and vision \cite{blip2,lyu2023macaw}. Regarding audio processing, LLM-based models have been studied for diverse tasks such as automatic speech recognition (ASR) \cite{Wu2023OnDA,Fathullah2023,slm,Yu2023ConnectingSE,llmDomainAdapt}, speech translation \cite{communication2023seamlessm4t,wang2023viola}, audio event detection and understanding \cite{gong2023listen}, etc. There are two paradigms for these models. The first one \cite{speechGPT,audioPaLM,wang2023viola} converts continuous audio into discrete tokens and then merges text and audio tokens into a shared vocabulary while the second one \cite{Fathullah2023,slm,Yu2023ConnectingSE} directly connects and adapts continuous representation of pretrained audio encoder into LLM. Although the first strategy is able to generate speech outputs from text or speech inputs by using codec-based discrete tokens, it may suffer from the information loss caused by discretization process hence might not be optimal for the speech-to-text task. In this work, we focus on the ASR task and we follow the second paradism. 

Due to the immense number of parameters in LLM models, it can be computationally impractical to adapt the whole LLM to a target task. Several approaches have been proposed to address this issue include: inserting adapter layers \cite{residualAdapter} or prefix embeddings \cite{prefixTuning} which are trained on target tasks. While these approaches are parameter-efficient, they increase the inference costs. Low-rank Adaptation (LoRa) \cite{hu2021lora} solves these issues by using low-rank matrices which are memory efficient during training and does not impact inference time.

\section{Proposed solution}
We first investigate more comprehensive fine-tuning schemes in Section 3.1. In Section 3.2, we introduce the proposed matching loss to improve alignment between modalities, followed by training and inference techniques to address the insertion issue in Section 3.3. 

\subsection{Fine-tuning schemes for LLM-based ASR}
Existing works only fine-tuned certain modules of LLM-based ASR models with limited configurations. For example, SLM~\cite{slm} only trained an adapter while keeping the other modules unchanged; whereas in \cite{Fathullah2023}, LoRa was used in the LLM and fine-tuned together with the adapter and entire encoder. It is also worth noting that existing works either fully fine-tuned the encoder or kept it frozen, ignoring the option for partial fine-tuning, which is more practical given the immense number of parameters in many current speech encoders. 

In this work, we investigate more thoughtful fine-tuning schemes with more options for each module. Specifically,
\begin{itemize}
\item  The LLM module typically has a very large size, thus it is impossible to fully fine-tune it using popular deep learning frameworks. Therefore, we consider freezing the LLM or fine-tuning it using a popular parameter-efficient method, e.g. LoRa \cite{hu2021lora}.

\item Foundation encoders such as HuBert \cite{hubert} or wav2vec2 \cite{baevski2020wav2vec} can be fully fine-tuned on recent advanced GPUs, such as the A100, which have high memory capacity. Hence, we consider three fine-tuning options for the encoder: frozen, partially fine-tuned with LoRa, and fully fine-tuned.

\item For the adapter, we use a fully connected layer adapter, which is similar to \cite{Yu2023ConnectingSE}. Specifically, it consists of a 1D convolution layer for subsampling, followed by the GeLu operation then a linear projection to have the same dimension as LLM embedding space. We denote this adapter as Conv1dMLP. Besides Conv1dMLP, we also consider different architectures for the adapter. Each of them requires a different number of parameters and has different expressive capacity. The first variant, denoted as DwsMLP, replaces a standard 1D convolution with a simple depthwise separable CNN, which requires fewer parameters. The second variant, denoted as Conv1dTransformer, uses Transformer \cite{transformer} layers after subsampling (similar to \cite{slm}), which generally have high expressive capacity but require substantially more parameters. 
\end{itemize}
\subsection{Improving modalities alignment by matching loss}
\label{subsection:matchingloss}
Past work \cite{Fathullah2023} has shown that the representation generated by the adapter is aligned, to some extent, with the LLM embeddings of the text. This alignment is implicitly achieved by fine-tuning the model using the ASR task, and there is no mechanism to explicitly force the two modalities to be aligned.

In this work, we propose a matching loss to explicitly align these modalities, which is not a trivial problem since the two sequences have different lengths. To overcome this challenge, we first apply cross attention, such as dot-product attention \cite{transformer}, between the text embedding and acoustic embedding sequences. This generates a sequence of acoustic representations that has the same length as the text embedding sequence. After that, we can easily apply standard loss functions such as mean square error (MSE) or Cosine distance between the text embedding and the newly generated acoustic representations. 

Formally, let us first denote $<$\textbf{X}, \textbf{Y}$>$ as a training utterance, where $\textbf{X}$ is a sequence of acoustic features and $\textbf{Y} = \{y_1, y_2,..., y_{|\textbf{Y}|} \}$ is a sequence of output text. Then, the matching loss, denoted as $L_m$, can be described as follows
\setlength{\belowdisplayskip}{0pt} \setlength{\belowdisplayshortskip}{0pt}
\setlength{\belowdisplayskip}{0pt} \setlength{\belowdisplayshortskip}{0pt}
\begin{align}
    \textbf{E}_{\textbf{Y}} &= Emb(y_1)\ Emb(y_2) ... Emb(y_{|\textbf{Y}|}) \label{embedding_fun} \\
    \textbf{X}_1 &= Adapt(Enc(\textbf{X})) \label{enc_fun} \\
    \textbf{H} &= CrossAtt(\textbf{E}_{\textbf{Y}}, \textbf{X}_1) = softmax(\frac{\textbf{E}_{\textbf{Y}} \textbf{X}_1^T }{\sqrt{d_{llm}}} ) \textbf{X}_1 \\
    L_m &= a\ MSE(\textbf{E}_{\textbf{Y}}, \textbf{H}) + b\  Cosine(\textbf{E}_{\textbf{Y}}, \textbf{H}) \label{loss_fun}
    \vspace{-0.3cm}
\end{align}
where $Emb$, $Enc$, $Adapt$ are LLM embedding, encoder and adapter functions respectively; $d_{llm}$ is the LLM hidden dimension; $a$ and $b$ are tunable hyper-parameters. The model is then trained using a combination of cross-entropy and $L_m$ losses.
\subsection{Inference and training methods to reduce insertion errors}
When analyzing model inference using the standard beam search algorithm, we found that models trained with a small amount of data, i.e. 10h, produced relatively high insertion errors. More importantly, when testing the robustness of our model on several out-of-domain test sets, we observed more severe insertion problems. Specifically, in these cases, models repeat an n-gram until reaching the output length limit. 

To alleviate insertion errors, one straightforward approach is to apply following constraints during beam search inference: 
\begin{itemize}
  \item Apply n-gram non-repetition constraint (denoted as NRNS): This constraint ensures that n-gram tokens of a specified length, e.g., 5, do not repeat during decoding. 
  \item Apply length penalty (denoted as LP): This method imposes a greater penalty on long decoding transcripts, thereby reducing insertion errors.  
\end{itemize}
Note that we utilize a publicly available LLM model from Hugging Face, where the above constraints are readily provided as options, namely, 'no\_repeat\_ngram\_size' and 'length\_penalty' \footnote{https://huggingface.co/docs/transformers/en/main\_classes/text\_generation}.

We observed that after applying inference constraints, insertion errors still persist substantially on out-of-domain test sets. We also noticed that the repetition problem frequently occurs in audio trunks containing only non-speech signals such as music or noise. We propose to apply the following training methods to address the insertion issue.
\begin{itemize}
  \item Data augmentation (referred to as DA): To enhance the system's robustness against variations in acoustic conditions and speaking styles, we perform data augmentation to introduce diverse acoustic conditions during training. Our data augmentation includes volume perturbation and speed perturbation, which are applied with certain probabilities. We refrain from adding random noise or music to training utterances, as we observed that it negatively impacts performance. 
  \item Fine-tuning a pre-trained model using non-speech segments with empty transcripts (referred to as NSET): Ideally, a model should not generate any output for audio trunks containing non-speech signals. We achieve this by augmenting the ASR training data with a non-speech corpus containing audio segments with empty transcripts. Subsequently, we fine-tune already trained models for a short duration. 
\end{itemize}



\section{Experimental setup}
\subsection{Dataset}
We conducted experiments using the LibriSpeech corpus \cite{librispeech}. In addition to training on the full 960h dataset, we trained models on two subsets: train-clean 100h and 10h. We report Word Error Rate (WER) results on the LibriSpeech dev-clean, dev-other, test-clean, and test-other sets. To assess model robustness, we also report results on two out-of-domain test sets: CoVoST2 \cite{wang2020covost} and GigaSpeech \cite{gigaSpeech}. For NSET training, we utilized the noise and music subsets of the Musan \cite{Snyder2015MUSANAM} corpus to construct the non-speech corpus. 
\subsection{Model settings}
We utilized the hubert-large-ll60k model \cite{hubert} (300M parameters) and the Vicuna vicuna-7b-v1.5 model \cite{vicuna2023} (7B parameters) from the Hugging Face website as the speech encoder and LLM, respectively. The encoder and LLM have hidden dimensions of 1024 and 4096, respectively. For the adapter, the 1D convolution has input and output channels of 1024 and 4096, respectively, and performs 8 times subsampling. The linear transformations in Conv1dMLP and DwsMLP have input and output dimensions of 4096. Consequently, Conv1dMLP has 48M parameters, while DwsMLP employs depthwise separable convolution, resulting in only 20M parameters. Conv1dTransformer, however, employs 2 layers of Transformer with a hidden dimension of 4096 and FFN dimension 2.5x larger, i.e., 10240, similar to \cite{slm}. Consequently, Conv1dTransformer has 320M parameters, significantly more than the other two adapter types.

For LoRa of the encoder, we implement $\{r=8, \alpha = 16\}$ on the query and value matrices of the self-attention module at each layer, resulting in 0.65M parameters. As for the LLM, we employ $\{r=16, \alpha = 16\}$ across all query, key, and value matrices of the self-attention, yielding 16M parameters.   

Table \ref{tab:list_config} presents all fine-tuning schemes, each corresponding to a specific configuration of each module. Due to the large number of configurations, we only explore adapter variants under two conditions: (1) with both encoder and decoder frozen, similar to \cite{slm}; (2) with both encoder and decoder fine-tuned using the LoRa method. 

\begin{table}[]\
\caption{Different fine-tuning configurations}
\fontsize{8pt}{8pt}\selectfont
\centering
\renewcommand\arraystretch{0.9}
\setlength{\tabcolsep}{2.4mm}
\begin{tabular}{llllr}
 \toprule
SID & Encoder & Adapter        & LLM & Params \\ 
\midrule
S1        & Frozen  & Conv1dMLP & Frozen  & 48M              \\ 
S2        & Frozen  & Conv1dMLP & LoRa    & 64M              \\ 
S3        & LoRa    & Conv1dMLP & Frozen  & 49M               \\ 
S4        & LoRa    & Conv1dMLP & LoRa    & 65M              \\ 
S5        & Full    & Conv1dMLP & Frozen  & 345M              \\ 
S6        & Full    & Conv1dMLP & LoRa    & 361M              \\ 
S7        & Frozen    & DwsMLP & Frozen    & 20M              \\ 
S8        & Frozen    & Conv1dTransformer & Frozen    & 320M              \\ 
S9        & LoRa    & DwsMLP & LoRa    & 37M              \\ 
S10        & LoRa    & Conv1dTransformer & LoRa    & 337M              \\ 
\bottomrule
\end{tabular}
\vspace{-0.5cm}
\label{tab:list_config}
\end{table}

\subsection{Training and inference setting}
We trained our models on A100 GPUs for 50k, 20k, and 10k steps for 960h, 100h, and 10h datasets, respectively. Checkpoints were saved at every 1k steps for the 960h and 100h data and every 500 steps for the 10h data. Following training, we selected five consecutive checkpoints with the best averaged validation loss and averaged them for evaluation. 

In our initial experiments on the matching loss, we found that the best results were achieved with \{a = 0.01, b = 0.04\}, hence we use this setting for all experiments in Section \ref{subsection:matchingloss}. 
For inference, we employed beam search with default settings from Hugging Face, i.e. \{beam\_size = 5, max\_length = 256, NRNS = 0, LP = 1.0\}
\begin{table*}[]
\vspace*{-0.3cm}
\begin{threeparttable}
\caption{WER results of different fine-tuning configurations (a) and matching loss (b)}
\fontsize{8pt}{8pt}\selectfont
\centering
\renewcommand\arraystretch{0.9}
\setlength{\tabcolsep}{1.3mm}
\begin{tabular}{l|rrrr|rrrr|rrrr}
\toprule
 & \multicolumn{4}{c|}{960h training data} & \multicolumn{4}{c|}{100h training data} & \multicolumn{4}{c}{10h training data} \\ 
\multirow{-2}{*}{SID} & dev-clean & dev-other & test-clean & test-other & dev-clean & dev-other & test-clean & test-other & dev-clean & dev-other & test-clean & test-other \\ 
 \midrule
S1 & 2.38 & 5.99 & 2.29 & 5.67 & 3.55 & 9.35 & 4.23 & 8.69 & 11.96 & 22.55 & 14.51 & 20.60 \\ 
S2 & 2.01 & 4.71 & 2.02 & 4.31 & 3.42 & 6.34 & 3.45 & 7.19 & 13.52 & 21.45 & 13.30 & 18.75 \\
S3 & 1.87 & 4.61 & 1.79 & 3.91 & 2.86 & 5.33 & 2.70 & 5.99 & 7.36 & 12.49 & 6.74 & 11.72 \\ 
S4 & 1.70 & 3.59 & 1.78 & 3.62 & 2.68 & 5.64 & 2.57 & 5.27 & 7.36 & 11.92 & 7.44 & 11.49 \\ 
S5 & 1.72 & 3.96 & 1.70 & 3.58 & 2.51 & 4.87 & 2.86 & 5.14 & 6.27 & 10.25 & 5.75 & 10.40 \\ 
S6 & 1.50 & 3.21 & 1.74 & 3.16 & 2.40 & 4.55 & 2.28 & 4.92 & 6.43 & 8.08 & 4.97 & 9.11 \\ 
S7 & 3.34 & 7.98 & 4.26 & 7.24 & 3.65 & 7.22 & 4.19 & 9.09 & 12.12 & 23.51 & 12.30 & 19.96 \\
S8 & 1.96 & 4.13 & 2.02 & 4.41 & 3.91 & 7.31 & 4.34 & 8.25 & 21.63 & 31.06 & 25.96 & 34.72 \\ 
S9 & 1.61 & 4.34 & 1.59 & 3.89 & 2.72 & 5.31 & 2.66 & 5.65 & 7.23 & 13.56 & 8.12 & 12.22 \\ 
S10 & 1.88 & 3.83 & 1.85 & 3.77 & 3.62 & 5.88 & 3.80 & 5.90 & 13.30 & 19.22 & 13.61 & 18.55 \\ 
 \bottomrule
\end{tabular}
\subcaption*{(a) Results of different fine-tuning configurations}
\begin{tabular}{l|rrrr|rrrr|rrrr}
\toprule
 & \multicolumn{4}{c|}{960h training data} & \multicolumn{4}{c|}{100h training data} & \multicolumn{4}{c}{10h training data} \\ 
\multirow{-2}{*}{SID} & dev-clean & dev-other & test-clean & test-other & dev-clean & dev-other & test-clean & test-other & dev-clean & dev-other & test-clean & test-other \\ 
 \midrule
S1 & 2.38 & 5.99 & 2.29 & 5.67 & 3.55 & 9.35 & 4.23 & 8.69 & 11.96 & 22.55 & 14.51 & 20.60 \\ 
T1 & 2.38 & 5.33 & 2.36 & 4.81 & 4.32 & 7.31 & 3.93 & 6.93 & 9.67 & 15.93 & 12.40 & 15.02 \\ 
S4 & 1.70 & 3.59 & 1.78 & 3.62 & 2.68 & 5.64 & 2.57 & 5.27 & 7.36 & 11.92 & 7.44 & 11.49 \\ 
T4 & 1.63 & 3.47 & 1.63 & 3.48 & 2.58 & 5.62 & 2.59 & 5.26 & 6.83 & 10.34 & 7.29 & 10.60 \\ 
\bottomrule
\end{tabular}
\subcaption*{(b) Results with and without matching loss. S1 and S4 are systems without matching loss while T1 and T4 are systems with matching loss}
\vspace{-0.5cm}
\label{tab:list_results_finetune}
\end{threeparttable}
\end{table*}
\section{Experimental results and analysis}
\subsection{Results of different fine-tuning scheme}
Table \ref{tab:list_results_finetune} (a) shows the results of different fine-tuning configurations which can be summarized as follows:
\begin{itemize}
  \item Applying LoRa to LLM significantly enhances performance across most test sets, especially on dev/test-other. For instance, on the 960h condition, S2 surpasses S1 by 21.4\%/24.0\% relative WER reduction (WER reduced from 5.99\%/5.67\% to 4.71\%/4.31\%) on dev/test-other and by 15.6\%/11.8\% (WER reduced from 2.38\%/2.29\% to 2.01\%/2.02\%) on dev/test-clean subsets. We posit that the dev/test-other subsets present challenging cases in terms of both acoustic and linguistic conditions, and fine-tuning the LLM enables better adaptation to the LibriSpeech domain.
\item In the encoder module, full fine-tuning yields the best outcomes, followed by partial fine-tuning with LoRa across all conditions. However, partial fine-tuning proves to be the most cost-effective. For instance, on 960h condition, partial fine-tuning S4 setting significantly outperforms the frozen setting S2 (WER reduced from 2.01\%/4.71\% to 1.70\%/3.59\% on dev-clean/other) while only adding 0.65M extra parameters. 
  \item Regarding adapter achitectures, the Conv1dTransformer clearly outperforms Conv1dMLP when both encoder and LLM are frozen and trained with 960h data, but does not exhibit benefit in other conditions. The other variant, DwsMLP, is generally worse than Conv1dMLP in most of conditions. 
\end{itemize}
In summary, fine-tuning both the encoder and LLM with LoRa and using Conv1dMLP as the adapter (S4 setting) achieves the optimal balance between performance and additional parameters. Thus, we employ S4 for subsequent experiments. Additionally, we utilize S1 for further experiments due to its crucial advantage: preserving all functionalities of the foundation encoder and LLM, which may be critical for certain applications.

\subsection{Results of matching loss}
Table \ref{tab:list_results_finetune}(b) presents the results of systems with and without matching loss. S1 and S4 are systems without matching loss taken from Table \ref{tab:list_results_finetune} (a), while T1 and T4 are corresponding systems with the matching loss. 
The matching loss enhances performance across most conditions, with a more pronounced improvement observed in S1 compared to S4. This discrepancy may stem from the LoRa modules in both encoder and LLM components, which already aid in aligning acoustic and text representations, thereby diminishing the benefit from the matching loss.
Notably, the improvement on dev/test-other subsets exceeds that of dev/test-clean. This discrepancy can be attributed to the presence of challenging cases in dev/test-other, characterized by higher modality mismatch, which the matching loss mitigates, resulting in improved performance. We then take the best model, i.e. T4, for experiments in the next section.

\subsection{Results of inference and training techniques}
\label{subsection:reduce_ins_results}

We analyze the impact of NRNS and LP constraints on T4 models trained with 960h (full data) and 10h (low resource) on dev sets. We observe a consistent trend in the results of both dev-clean and dev-other, so we present only the results for dev-other in Table \ref{tab:ngram_constraint}. We also report the insertion error rate (IER) in some cases for analysis. Each constraint helps to reduce IER in the 10h conditions. For instance, NRNS = 10 reduces IER from 1.88\% to 0.89\% on dev-other, resulting in a WER improvement from 10.35\% to 9.35\%. However, combining these constraints does not yield additional benefits, suggesting that LP and NRNS may not be complementary. Overall, the optimal setting is NRNS = 10, LP = 0. 

\begin{table}[]
\begin{threeparttable}
\caption{Effect of NRNS and LP on WER and IER of dev-other. IER are showed in parentheses}
\vspace*{-\baselineskip}
\fontsize{8pt}{8pt}\selectfont
\centering
\renewcommand\arraystretch{0.9}
\setlength{\tabcolsep}{2.4mm}
\begin{tabular}{lcccc}
\toprule
Data & 0            & 5    & 10          & 15   \\ 
\midrule
960h & 3.47 (0.33)  & 3.46 & 3.44 (0.33) & 3.44 \\ 
10h  & 10.34 (1.88) & 9.36 & 9.35 (0.89) & 9.46 \\ 
\bottomrule
\end{tabular}
\subcaption*{(a) Effect of NRNS constraint on WER and IER}
\begin{tabular}{lccccc}
\toprule
Data              & NRNS & 1.0 & 0.5 & 0 & -0.5 \\ 
\midrule
\multirow{2}{*}{960h} & 0    &  3.47 (0.33)   &  3.49 (0.32)  & 3.50  &   3.50   \\ 
                      & 10   &   3.44 (0.33)  &  3.46 (0.33)  &  3.47 &   3.49   \\ 
\midrule
\multirow{2}{*}{10h}  & 0    &  10.34 (1.88)   &  9.44 (0.94)   &  9.61  &   9.71   \\ 
                      & 10   &   9.35 (0.89)  &  9.35 (0.89)   & 9.51  &  9.61    \\
\bottomrule
\end{tabular}
\subcaption*{(b) Effect of the LP constraint on WER and IER}
\vspace{-0.3cm}
\label{tab:ngram_constraint}
\end{threeparttable}
\end{table} 

We assess the above constraints on test-clean, test-other, and two out-of-domain test sets. Results are summarized in Table \ref{tab:contraints_testsets}. Applying constraints reduces IER and consequently enhances performance on in-domain test sets under the 10h training condition. Importantly, applying constraints significantly reduces IER on out-of-domain test sets for both 960h and 10h conditions. For instance, for the model trained with 960h, IER decreases from 33.06\% to 14.52\% in GigaSpeech, leading to a WER reduction from 46.42\% to 27.83\%. 
\begin{table}[]
\caption{WER and IER (showed in parentheses) on different test sets without and with contraints during inference.}
\fontsize{8pt}{8pt}\selectfont
\centering
\renewcommand\arraystretch{0.9}
\setlength{\tabcolsep}{0.5mm}
\begin{tabular}{lccccc}
\toprule
Data                  & Constraint       & test-clean  & test-other   & CoVoST2       & GigaSpeech    \\ 
 \midrule
\multirow{2}{*}{960h} & No & 1.63 (0.15) & 3.48 (0.40)  & 31.47 (16.94) & 46.42 (33.06) \\
                      & Yes   & 1.63 (0.15) & 3.49 (0.40)  & 19.73 (5.21)   & 27.83 (14.52)  \\
 \midrule
\multirow{2}{*}{10h}  & No & 7.29 (1.51) & 10.60 (1.83) & 47.89 (24.80) & 59.67 (39.02) \\ 
                      & Yes   & 6.37 (0.61) & 9.73 (1.00)  & 33.09 (10.07) & 44.39 (23.86) \\ 
\bottomrule
\end{tabular}
\vspace{-0.3cm}
\label{tab:contraints_testsets}
\end{table}

We examine the effects of training methods: DA and NSET, summarized in Table \ref{tab:results_training}. DA significantly improves performance across all conditions, notably in low-resource training and out-of-domain evaluation scenarios. Notably, in scenarios like 960h training and testing on GigaSpeech, the reduction in IER is close to that of WER, indicating that most of the WER benefit stems from reduced insertion errors. Moreover, DA not only reduces insertion errors but also leads to fewer deletion/substitution errors in test-clean and test-other, particularly in the 10h condition. NSET training remarkably reduces insertion errors in out-of-domain test sets. However, this method sometime increases deletion errors resulting in higher WER, particularly in the 10h training condition. Although the combination of DA and NSET does not always yield the lowest insertion errors, it generally results in better WER due to fewer deletion errors. Overall, DA+NSET demonstrates increased robustness while effectively suppressing insertion errors, as demonstrated in an example from CoVoST2 corpus in Table \ref{tab:example}.

\begin{table}[]
\fontsize{8pt}{8pt}\selectfont
\centering
\renewcommand\arraystretch{0.9}
\setlength{\tabcolsep}{0.4mm}
\caption{WER and IER (showed in parentheses) on different test sets with different training methods}
\begin{tabular}{lccccc}
\toprule
Data                  & Methods & test-clean  & Test-other   & CoVoST2 & GigaSpeech       \\ 
\midrule
\multirow{4}{*}{960h} & default                & 1.63 (0.15) & 3.49 (0.40)         &    19.73 (5.21) & 27.83 (14.52)         \\ 
                      & DA  & 1.61 (0.15)        & 3.37 (0.39)        &  18.93 (4.67)  &    23.90 (10.81)     \\ 
                      &   NSET   & 1.70 (0.16)         & 3.53 (0.39)               & 19.05 (3.77)    &   16.78 (3.22)     \\  
                      & DA+NSET&     1.67 (0.15)        &   3.48 (0.41)       &  18.40 (3.79)     &    16.58 (3.33)     \\  
\midrule
\multirow{4}{*}{10h}  & default                & 6.37 (0.61) & 9.73 (1.00) &  33.09 (10.07) & 44.39 (23.86) \\  
                      & DA &    5.36 (0.46)         &       8.65 (0.83)      &       31.25 (9.20)  &  40.73 (21.51)        \\  
                      & NSET &    7.05 (0.60)         &     10.57 (1.06)       &    32.84 (8.44)     &    26.36 (5.56)   \\  
                      & DA+NSET&      5.80 (0.48)        &  9.13 (0.94)          &     31.40 (8.02)            &  26.87 (6.84)    \\
\bottomrule
\end{tabular}
\vspace{-0.1cm}
\label{tab:results_training}
\end{table}

\begin{table}[]
\caption{Decoding hypotheses generated by models trained with different methods for the utterance common\_voice\_en\_37177}
\fontsize{8pt}{8pt}\selectfont
\centering
\renewcommand\arraystretch{0.9}
\begin{tabular}{ll}
\toprule
ref / hyps & Text \\ 
\midrule
reference & ican't stand the sight of blood \\ \hline
hyp - default & \begin{tabular}[c]{@{}l@{}}or ican't stand the sight of blood or ican't \\ stand the site of blood ... or ican't stand\\  the sight of blood\end{tabular} \\ 
hyp - DA & ican't stand the sight of blood i \\ 
hyp - NSET & \textless{}empty\textgreater{} \\ 
hyp - DA+NSET & ican't stand the sight of blood \\ 
\bottomrule
\end{tabular}
\vspace{-0.3cm}
\label{tab:example}
\end{table}

\section{Conclusion}

We presented a systematic approach to integrate a speech encoder with an LLM for the ASR task. Our findings indicate that employing the parameter-efficient LoRa method on both the encoder and LLM represents the most cost-effective fine-tuning strategy. Additionally, we introduced a matching loss to enhance alignment between modalities, thereby improving ASR performance. Moreover, we explored various inference methods alongside training techniques to mitigate insertion errors. Results demonstrate a substantial reduction in insertion errors using these methods, resulting in significantly improved ASR performance, especially for out-of-domain test sets.

\bibliographystyle{IEEEtran}
\bibliography{main}

\end{document}